\documentclass{article}
\usepackage{CJKutf8}
\usepackage{graphicx} 
\usepackage[T1]{fontenc}
\usepackage{ipa}
\usepackage[utf8]{inputenc}
\usepackage{amsmath}
\usepackage{appendix}
\usepackage{booktabs} 
\usepackage{float}
\usepackage{multicol}
\usepackage{multirow}
\graphicspath{{images/}}
\usepackage{xcolor}
\usepackage[export]{adjustbox}
\usepackage{natbib}
\usepackage{longtable}
\usepackage{color}
\usepackage{tipa}
\usepackage{geometry}
\usepackage{hyperref}
\usepackage{comment}
\usepackage{authblk}
\usepackage{textcomp}
\usepackage{subcaption}
\usepackage{comment}
\usepackage{rotating}

\title{Word meaning co-determines vowel-inherent spectral change. A corpus-based investigation of conversational Mandarin.}

\author[a]{Xiaoyun Jin}
\author[b]{Mirjam Ernestus}
\author[c]{R. Harald Baayen\thanks{Funding: This work was supported by the European Research Council under Grant SUBLIMINAL (\#101054902) awarded to R. Harald Baayen}}

\affil[a]{Quantitative Linguistics, Eberhard Karls Universität Tübingen, 72074 Tübingen, Germany\\Email: xiaoyun.jin@uni-tuebingen.de}

\affil[b]{Center for Language Studies, Radboud University, 6525 HT Nijmegen, The Netherlands\\ Email: mirjam.ernestus@ru.nl}

\affil[c]{Quantitative Linguistics, Eberhard Karls Universität Tübingen, 72074 Tübingen, Germany\\ Email: harald.baayen@uni-tuebingen.de}

\date{}

\begin{document}
\begin{CJK}{UTF8}{bsmi} 
\begin{CJK}{UTF8}{gbsn} 

\maketitle


\newpage

\begin{abstract}

\noindent 
This study investigates vowel-inherent spectral change (VISC) in spontaneous conversational Mandarin. Using the generalized additive model and word embeddings from distributional semantics, we show that, when controlling for variables such as vowel duration, gender, speaker identity, co-articulation, vowel identity, and utterance position, vowel formant trajectory dynamics have word-specific components that are tied to their meaning in context:  The F1 and F2 trajectories of words can be predicted from their contextualized embeddings with an accuracy that substantially exceeds a permutation baseline.  Challenging modular cognitive models of speech production, these results indicate that, words' semantics co-determine the fine details of their articulation. \\ \ \\




\noindent
\textbf{Keywords: vowel formants, distributional semantics, conversational speech, Mandarin,
Discriminative Lexicon Model} 
\end{abstract}

\newpage

\section{Introduction}
Vowel quality is primarily characterized by the spectral properties of speech. Formant frequencies, specifically the first formant (F1) and the second formant (F2), constitute the primary acoustics cues of vowel quality in human speech \citep{ladgfoged2001course}.  F1 is inversely related to vowel height while F2 is inversely related to vowel backness, reflecting the vertical and front-back position of the tongue respectively, albeit with speaker- and context-dependent variation \citep{hillenbrand1995acoustic,howie1976acoustical}.  These two formant parameters play a foundational role in vowel production and perception. Distinct vowel categories are systematically modulated by F1 and F2 values in production \citep{peterson1952control}. In perception, F1 and F2 are primary cues for vowel identification, and small changes in frequency values leads to misidentification \citep{neel2008vowel}.

\subsection{Vowel inherent spectral change}
Traditional approaches to vowel characterization have relied on static formant measurements. Typically, formant values are extracted from a single time point such as the vowel midpoint or averaged values across the entire vowel duration. Under this statistic perspective, monophthongs have been described as relatively steady \citep[e.g.,][]{peterson1952control}. In contrast, diphthongs generally involve systematic formant movement over time between two vocalic targets \citep{holbrook1962diphthong,stevens2000acoustic}. However, a growing body of research \citep[e.g.,][]{nearey1986modeling} has clarified that the center stable state of monophthongs is also subject to fluctuations, referred to as Vowel Inherent Spectral Change (VISC). VISC has often been measured by inspecting formant values at a small number of key points across the entire vowel.  For instance, \citet{nearey1986modeling} measured the formants values at three points for lab-recorded English vowels and observed substantial formant movement in monophthongs. Using similar methods, other studies addressed VISC for British English and for standard Dutch \citep{williams2015beyond}, for Australian English \citep{elvin2016dynamic}, and for Spanish \citep{morrison2009l1} also observed dynamic spectral formant change. These studies collected point measures of formant trajectories in lab-recorded speech, such as the difference between the beginning and end of the ``steady state'' (20\% into the vowel, and 80\% into the vowel) possibly with 1 up to 5 intermediate measurement points.  \cite{morrison2007testing} suggested that other measures, such as \textit{onset + slope} and \textit{onset + direction} might be able to better capture the nature of formant change. 

Research on VISC has been carried out predominantly on laboratory-controlled speech. Laboratory speech offers high acoustic quality and tight experimental control. On the other hand, laboratory speech can be very different from natural spontaneous speech. More recent studies made use of  Generalized Additive Mixed Models \citep[GAMM][]{Wood:2017} to analyse formant trajectories in connected speech or spontaneous speech. GAMMs are particularly well-suited for modelling non-linear time series data, as they can model entire trajectories while at the same time taking into account the variation contributed by random-effect factors such as speaker, and control predictors such as duration.  For example, \cite{renwick2020modeling} studied the formant trajectories based on a semi-spontaneous speech corpus of American English speakers, by extracting F1 and F2 trajectories at five time points of the target tokens.  The results of modelling with GAMMs indicated greater vowel-inherent raising of /i/ among African American speakers. \cite{brandt2021dynamic} analyzed recordings of trained and experienced German broadcast announcers and extracted the full formant trajectories of their target words.   They found differences in the formant movements between vowels  that depended on the predictability of the context in which they occurred. \cite{yuan2013spectral} investigated VISC in Mandarin using a corpus of monosyllabic words produced in isolation. This study parameterized MFCCs at three points of the vowels and observed vowel-inherent spectral change primarily in the first half of the vowel. The also observed that the rate of spectral change correlated with vowel duration. 

\subsection{Phonetic and Semantic Effects on formants}

The way in which formants are realized is co-determined by many different factors. It is well-established that vowels' first and second formants are co-determined by the place of articulation of adjacent consonants \citep{Ladefoged:1975,hillenbrand2001effects}. These changes typically affect only part of the vowel, for Consonant-Vowel-Consonant (CVC) syllables resulting in patterns with an initial rising or falling contour, followed by a stable state, followed in turn by a final falling or rising contour.  Co-articulation with vowels in adjacent syllables can also affect formant realization \citep{cole2010unmasking,zellou2018gradient,viswanathan2014information}. For example, in a Mandarin sequence such as /i/ – stop – /a/ (like /b{i}2-b{a}4/), the onset F2 of the vowel /a/ is significantly higher than when /a/ is produced in isolation. This reflects carry-over (right‑to‑left) coarticulation from the high‑front vowel /i/ \citep{wang2012study}.

F0 (fundamental frequency) can influence the perception and discrimination of formant frequencies, particularly F1 and F2. The interaction between F0 and formants, where changes in F0 have been shown to shift perceptual boundaries or affect the discriminability of closely spaced vowels\citep{chladkova2009line,shaw2016influences,gardner1989perceptual}.

Vowel duration and speech rate have been found to influence the realization of vowel formants \citep{gendrot2005impact,pitermann2000effect}. In particular, F1 and F2 values of short vowels diverge more from the references values due to reduction \citep[see also][]{ernestus2014acoustic}. Position in the utterance also affects the realization of formants.  Especially in utterance-final position, formant transitions have been observed to be shallower, but this effect is not consistent across studies  \citep{kuo2014formant}.  In addition, the formants of female speakers tend to have higher frequencies than the formants of male speakers \citep{coleman1971male}. 

In addition to the above-mentioned prosodic and sociolinguistic variables, a novel factor co-determining phonetic realization that is emerging from recent investigations is semantics.  
For English, \citet{plag2015homophony} reported that the duration of English word-final [s] co-varies with the morpho-syntactic function that this segment is realizing. \citet{Gahl_Baayen_2024} reported that the spoken word duration of English homophones is in part predictable from the meanings of these homophones. Similar results have been obtained for Japanese \citep{saito2026sound}. Furthermore, a series of studies of the phonetic realization of tones in conversational Mandarin documented effects of a word's meaning on the phonetic realization of pitch contours in spontaneous conversational  speech, independently of a wide range prosodic factors \citep{Chuang_Bell_Tseng_Baayen_2026,lu2025realization,JIN2026101495}. 
These studies also clarified that the way in which words' pitch contours vary can be predicted above chance level from the contextualized embeddings of these words, supporting the possibility that words' meaning in context is a co-determinant of the realization of their pitch contours. 


\subsection{The present study}

In naturalistic and spontaneous speech, the way in which formants change over time can deviate substantially from the steady-state patterns typically observed in careful laboratory speech.
In the present study, we consider the question of whether the formant trajectories of Mandarin monophthongs in spontaneous speech have word-specific components that are tied to the meanings of these words in context.  More specifically, our research goals are to find answers to the following questions.

\begin{enumerate}
    \item \textbf{Is vowel inherent spectral change present not only in laboratory-controlled speech \citep{yuan2013spectral}, but also in conversational spoken Mandarin?} While VISC has been robustly documented in lab-controlled speech \citep{yuan2013spectral}, its presence in spontaneous Mandarin remains an open question. Confirming VISC in conversational Mandarin would extend the scope  of this phenomenon.  
    \item \textbf{Do words' VISC contours have word-specific components that are independent of prosodic and sociolinguistic factors?} In this study, we will exploit the potential of the generalized additive model \citep[GAM,][]{Wood:2017,wieling2018analyzing} for understanding the formant trajectories of vowels in conversational speech. What makes the generalized additive model especially attractive is GAMs decompose an observed formant trajectory into a set of additive component trajectories that are tied to linguistic variables. GAMs make it possible to tease apart the contributions of prosodic and sociolinguistic factors from word-specific effects, and to detect the presence of word-specific VISC contours.
    \item \textbf{Can word-specific VISC components be predicted above chance from contextualized embeddings, which would provide evidence for VISCs being co-determined by words' meanings?} If word-specific VISC components can indeed be detected, a further question is whether embeddings from distributional semantics can predict these words-specific VISCs.  If these embeddings predict the word-specific VISC patterns beyond prosodic and sociolinguistic controls, that would support the hypothesis that words' spectral dynamics are not arbitrary but are systematically linked to words' semantics. 
\end{enumerate}

\noindent
The remainder of this paper is organized as follows. In Section~\ref{sec: data}, we provide details about the corpus, the extraction and processing of formant trajectories, the covariates considered, and the statistical analysis methods employed. Section~\ref{sec: results} presents the results of our analyses. Finally, Section~\ref{sec: gendisc} discusses the broader implications of our findings.

\section{Method}\label{sec: data}

\subsection{Data}
The formant data were extracted from the Taiwan Mandarin Spontaneous Corpus \citep{fon2004preliminary}, comprising 55 casual conversations. Each word in this corpus is transcribed using traditional Chinese characters. Furthermore, the corpus provides detailed information on words' phones and their alignment with the audio signal. In the corpus, 118,592 single-character (and single-syllable) words are attested, the vowels of which are transcribed with 16 different vowels, 12 monophthongs and 4 diphthongs. For this study, the dataset was restricted to monosyllabic words with the vowels /a, i, u, \textschwa/ with as further restrictions the exclusion of coda consonants and pre-nuclear glides.  This resulted in a dataset of 53,139 tokens representing 699 word types.  Around 85\% of the tokens belong to 30 high-frequency words. To prevent statistical analyses from being heavily biased in favour of these high-frequency words, for words with token frequencies exceeding 220, we randomly sampled 220 tokens across all 55 speakers. Furthermore, word types with fewer than ten tokens were excluded from the dataset.  The total number of word tokens in our dataset is 8,185, representing 103 types.

The Montreal Forced Aligner (MFA) \citep{mcauliffe2017montreal} was used to determine the boundaries between consonants and vowels. Among the 8,185 tokens, 404 tokens were not assigned a boundary, typically due to being too short. To assess the quality of the alignment, the MFA-aligned tokens were audited manually. 301 tokens were excluded due to miss-match between the speech sounds and their corresponding words when extracted in isolation,  a phenomenon that is quite common in human perception of spontaneous speech \citep[e.g.,][]{Ernestus:Baayen:Schreuder:2002}.

Among the remaining 7,480 tokens,  to ensure the length and quality of the audio files, the spectrogram of each token were inspected manually in Praat \citep{boersma2001speak}. A total of 119 tokens with extreme short durations and 345 tokens with noisy spectrograms were excluded from the analysis, which left us a dataset of 7,016 tokens.

\begin{figure}
    \centering
    \includegraphics[width=0.8\linewidth]{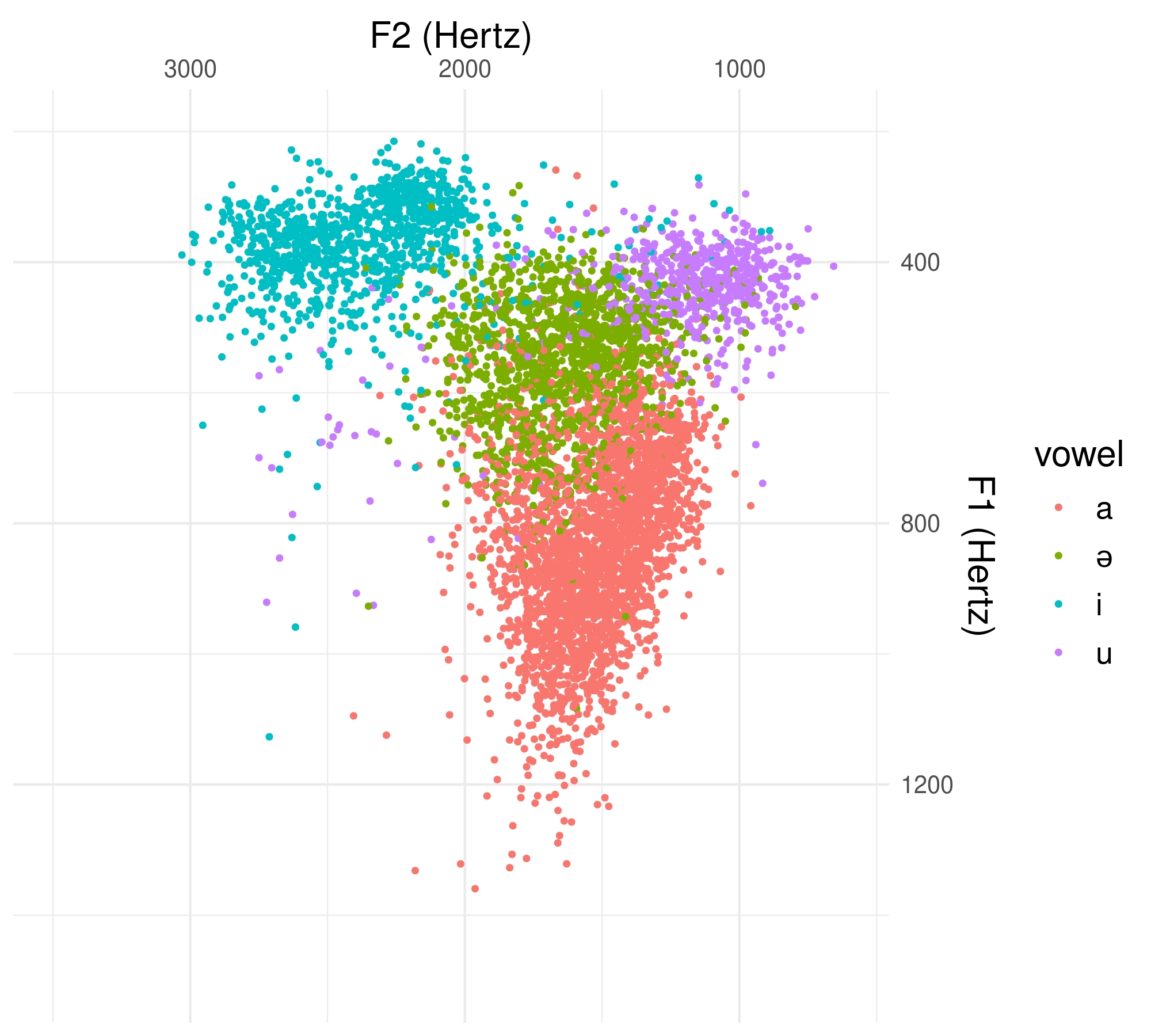}
    \caption{Distribution of four Mandarin vowels F1-F2 formant space. Each dot represents the mean formant value for a word token.}
    \label{fig:mean vowel}
\end{figure}

For the next step, formant values were measured using linear predictive coding (LPC) implemented in Praat \citep{boersma2001speak}.\footnote{Difficulty in extracting reliable formant values from spontaneous speech has been documented in previous research \citep{kiefte2017modeling,rathcke2017beauty,houzar2022intra,dicanio2015vowel,fulop2007s}. Changing speech rate, inevitable background noises and drastic speech style changes caused by mood shifts such as laughter can significantly degrade the quality of the acoustic signal and introduce considerable inconsistency in formant value estimates. LPC, the conventional formant tracking algorithm, often struggles to extract consistent trajectories under natural speech conditions\citep{shadle2016comparing,van2012predicting}. To address these issues, we also attempted to apply the formant trajectory extraction proposed by \cite{wempe2018sound}. However, the formant range within one token was found to be larger than that obtained using the LPC algorithm.  Although the values extracted using LPC also exhibited some extremely large within-token variability, overall the distribution of the mean formant value of each token  remained within an acceptable range, see Figure~\ref{fig:mean vowel}. Therefore, in subsequent analyses, we adopted the formant values derived from the LPC algorithm.} Analyses used a window length of 25 ms, a time step of 2.5 ms, pre-emphasis from 50 Hz, and a maximum formant frequency of 5500 Hz. Tokens with fewer than five measurement points were excluded, which left us with a dataset of 6,252 tokens representing 102 word types. 

For each of the 6,252 word tokens, we estimated their meaning in context, henceforth sense, using the word sense disambiguation system proposed by \citet{hsieh2024resolvingregularpolysemynamed}, which makes use of the Chinese WordNet \citep{huang2010constructing} in combination with a GPT-2 language model developed by the Academia Sinica in Taiwan for Mandarin Chinese \url{https://github.com/ckiplab/ckip-transformers}. Across the 102 word types in the dataset, a total of 392 distinct senses were identified. More than 70\% of the word types exhibited polysemy. To ensure sufficient data for model fitting, only senses represented by more than 5 tokens were included. The resulting sense-annotated dataset comprised 5,828 tokens from 87 word types with in all 188 word senses.   Table~\ref{tab:vowel_exmaple} provides, for each of the four vowels under investigation, the number of word types, the number of word tokens and an example word. Any given vowel is found across multiple words, and is realized by multiple speakers. On average, each speaker produced about 38.1 different word types (median 37).  For words with longer durations, more formant measurements were extracted. The range of vowel duration in the dataset is 8 ms to 824 ms (mean: 148 ms and median: 108 ms).  On average, vowel formants were represented by 39.8 measurement points per token. Figure~\ref{fig:mean vowel} presents the mean formant values for each of the tokens in the F1-F2 plane.

\begin{CJK}{UTF8}{bsmi}

\begin{table}[htpb]
    \centering
    \resizebox{0.6\linewidth}{!}{
    \vspace*{0.5\baselineskip}
\begin{tabular}{llll} \hline
    Vowel & Types &  Tokens &  Mandarin Example  \\
    \hline
   i &  28 & 1087 & \emph{\textit{你}} (`you', n\v{\i}) \\ \hline
   u &  13 & 593 & \emph{\textit{組}} (`group', z\v{u})\\ \hline
   \textschwa &  20 & 1415 & \emph{\textit{和}} (`and', h\'{e})\\ \hline
   a & 27 & 2733 & \emph{\textit{打}} (`beat', d\v{a})  \\\hline   
\end{tabular}
    }
    \caption{Types and tokens of words with one of the four vowels under investigation, in the sense-annotated dataset. For each vowel, an example Mandarin word and its corresponding English translation and Pinyin is presented.}
    \label{tab:vowel_exmaple}
\end{table}
\end{CJK}

\subsection{Predictors}

\subsubsection{Core predictors}
\textbf{Place of articulation.} Results from previous studies \citep[e.g.,][]{farnetani1993anticipatory,hillenbrand2001effects} have shown that \texttt{place of articulation} of the preceding consonant is a strong determinant of formant trajectories. Particularly large shifts in formant patterns have been observed for rounded vowels in alveolar environments \citep{stevens1963perturbation,bouguyon2020coarticulatory}. Therefore we included \texttt{place of articulation} as a factorial predictor. In our dataset, seven categories of \texttt{place of articulation} are distinguished, as shown in Table~\ref{tab:poa_distribution}. For tokens without initial consonant, place of articulation is coded as NULL.  As the effect of \texttt{place of articulation} is expected to vary depending on the target vowel, we created a factor, \texttt{place of articulation$\times$vowel}, with a total of 16 levels, each specifying one of the combinations of vowel and place of articulation that are attested in our dataset. 

In our dataset, only Mandarin words with CV structure are selected. When \texttt{place of articula\-tion$\times$vowel} is included as a predictor, this factor takes the mechanical articulatory effects on formants into account. However, this factor is to a considerable extent confounded with word identity: there are 87 word types, 7 levels for place of articulation, and hence on average only 12.4 different word types for each level of \texttt{place of articulation by vowel} (cf. Table~\ref{tab:poa_distribution}).  
Therefore, this predictor and word will not be considered jointly in the same statistical analysis. 

An effect of onset consonant, rather than place of articulation, on formant has been reported in \cite{kiefte2017modeling}.  However, the combination of the initial consonant and the vowel is almost completely confounded with word, with a few homophones and tone-differentiated homographs as exceptions. Therefore, in the current study, we do not include the identity of the onset consonant as a predictor. Instead, we include \texttt{word} as a core predictor.

\begin{table}[htb]
    \centering
    \begin{tabular}{ccccc}
         place of articulation & consonant & vowel & number of tokens & number of word types\\ \hline
        alveolo-palatal & /\textctc, t\textctc \super h, t\textctc/ &/i/ & 349 & 13\\
        alveolar & /t, l, n, ts, t\super h, ts\super h/ & /a, u, i, \textschwa/ & 1561;132;436;711 & 32\\
        bilabial& /p, p\super h, m/ & /a, u, i, \textschwa/ & 882;141;76;7 &14 \\
        labiodental & /f/ & /a/ & 5 &1 \\
        NULL&   & /a, i, \textschwa/ & 157;226;8 &4 \\
        retroflex& /\textctz, t\textrtails\super h, \textrtails, t\textrtails/ & /a, u, \textschwa/ & 128;307;279 &14\\
        velar& /k, k\super h, h/ & /u, \textschwa/& 13;410 &9\\
    \end{tabular}
    \caption{Overview of \texttt{place of articulation} and its corresponding consonants and vowels and number of tokens.}
    \label{tab:poa_distribution}
\end{table}

\noindent
\textbf{Word.} Defined by its orthographic form in the corpus, 
\texttt{word}, is our second core predictor, with 87 levels.

\noindent
\textbf{Word sense.} The meaning of a monosyllabic Mandarin word in isolation is often ambiguous \citep{huang2002nature}. In our dataset of 5,828 tokens across 87 word types, 188 different senses are attested. For example, for  啊 \textit{a0}, the following senses are distinguished:

\begin{CJK}{UTF8}{bsmi}
\begin{center}
\begin{tabular}{ll} 

表解釋或提醒對方的語氣 &  `to explain or remind'  \\
表反問的語氣         &  `questioning' \\ 
表列舉省略的語氣      &  `enumerating'  \\
表感嘆的語氣         &  `exclamatory'  \\
表停頓的語氣         &  `deliberate pause' \\

\end{tabular}
\end{center}
\end{CJK}

\subsubsection{Control variables}

\textbf{Gender} was included as factorial predictor as female speakers tend to exhibit higher formant frequencies compared to male speakers \citep [e.g.,][]{coleman1971male}. 

\noindent
\textbf{Vowel} is a factor with four levels (/a,i,u,\textschwa/). This factor accounts for major differences in the distributions of F1 and F2 values (cf. Figure~\ref{fig:mean vowel}). 


\noindent
\textbf{Time} was normalized to the interval [0,1]. Since formants were measured every 5 ms, vowels with longer duration have more measurement points in the [0, 1] interval. 

\noindent 
\textbf{F0} (fundamental frequency) can influence the perception and discrimination of formant frequencies, particularly the first (F1) and second (F2) formants \citep{chladkova2009line,shaw2016influences}. Higher F0 values significantly reduced listeners' ability to discriminate small changes in vowel formant frequencies  \citep{kewley1996fundamental}. 
In the current study, log-transformed F0 was used.  As the effect of \texttt{logF0} may play out differently over time, we used a tensor product smooth to model the interaction of \texttt{time} and \texttt{logF0}. 
%

\noindent
\textbf{Duration} has been found to influence the realization of vowel formants \citep{hillenbrand2000some, gendrot2005impact}. We included log-transformed vowel \texttt{duration} as a predictor, using the logarithmic transformation in order to avoid adverse effects of outliers.

\noindent
\textbf{Speaker} was included as random-effect factor in order to accommodate potential speaker-specific formant values.  

\noindent
\textbf{Utterance position.} In American English, the realization of formant trajectories can vary with a word's  position within an utterance \citep{kuo2014formant}, due to changes in intonation and changes in speech rate such as phrase-final lengthening.  Whether formants in colloquial Mandarin are likewise affected has, to our knowledge, not been researched.  We calculated the normalised position of a target word in its utterance, henceforth \texttt{utterance position}, and included it as a covariate in the statistical analysis. An utterance was defined as a sequence of words preceded and followed by a perceivable pause (regardless of its duration), following the annotation provided by the corpus. The normalised \texttt{utterance position} of a given word token in an utterance is the position at which the token occurs divided by the total number of words in the utterance. Hence, this predictor is bounded between 0 and 1. 

\noindent
\textbf{Vowel sequence.}  In the presence of an adjacent vowel, vowels can undergo co-articulation \citep{cole2010unmasking,zellou2018gradient}.For our analyses, we looked up in the corpus the vowel of the preceding and following syllable. If the adjacent vowel is a diphthong, then only the adjacent part of the diphthong is included. When the target token occurs next to a pause, which means the preceding/following vowel is non-existent, then it is coded as \texttt{NULL}. The preceding and following vowels were coded for their height. We distinguished between three vowel heights: `high', `mid' and `low'. In total, 7 `high' vowels (/i/, /u/, /y/, /\textbaru/, /\textupsilon/, /\textsci/, /\textbari/,), 4 `mid' vowels (/\textschwa/, /e/, /\textepsilon/, /\textrhookschwa/) and 2 `low' adjacent vowels (/\textscripta/, /a/) are attested in our dataset.  Given 3 heights and NULL, the total number of possible combinations of preceding and following vowel height for any given vowel is 16. As the effect of neighbouring vowels is expected to vary with target vowel, we defined a random-effect factor, \texttt{vowel sequence} with in theory 4 $\times 16 = 64$ levels, representing  all possible combinations of adjacent vowel heights and target vowel. However, only 48 of these combinations are attested in the dataset.

\noindent 
\textbf{Bigram probabilities.} The forward and backward bigram probabilities of words were defined following  \citep{Gahl_Baayen_2024,Chuang_Bell_Tseng_Baayen_2026,lu2024form}, as probability or information content in context is known to codetermine phonetic realization \citep[see, e.g.,][]{van2003efficient,ernestus2000voice}. As the effects of the \texttt{bigram probability} measures did not enhance model quality, they will not be reported in the analyses below. For detailed discussion, see section~\ref{appendix: bigram} in the Appendix.

\subsection{Statistical Modelling}

For the analysis of F1 and F2 values as a function of (normalized) time, we used the generalized additive model (GAM), as implemented in the \textbf{mgvc} package \citep{wood2015package} for R \citep{r2013r}. Formant values were log-transformed. We fitted separate GAM models for F1 and F2, including in each analysis the data of all four vowels.  As there is substantial autocorrelation in the time series of formant values,  we included an AR(1) model for the residuals, with $\rho = 0.85$ for F1 and $\rho = 0.9$ for F2. We constrained the number of basis function to 4, mainly to avoid having highly complex parallel smooths for time.

For each of the formants, we built a baseline model incorporating all control predictors and included one core predictor, represented below as $X$. This core predictor varied across models:  \texttt{place of articulation$\times$vowel}, \texttt{word}, or \texttt{word sense}. These three predictors were assessed in separate models in order to avoid collinearity and concurvity.  We used Akaike’s Information Criterion \citep[AIC;][]{Akaike:1974} to assess the relative importance of the predictors.

The baseline models for F1 and F2 are nearly identical, and differ only with respect to two terms. 

\parbox{0.48\textwidth}{
\begin{tabbing}
mm \= \texttt{pitch(logF1)} \= $\sim$ \= \texttt{gender} + \texttt{vowel} \kill
    \> \texttt{pitch(logF1)} \> $\sim$ \> \texttt{gender} +  \texttt{vowel} +\\
    \> \>\> s(\texttt{logF0}, by = vowel, k = 4) +  \\
    \> \>\> ti(\texttt{normalized\_time}, \texttt{logF0}) + \\
    \> \>\> s(\texttt{duration}, by = vowel, k = 4) + \\
    \> \>\> s(\texttt{normalized\_time}, \texttt{speaker}, bs = ``fs", m=1) + \\
    \> \>\> s(\texttt{utterance position}, by = vowel, k = 4) + \\
    \> \>\> s(\texttt{normalized\_time}, \texttt{vowel sequence} , bs = ``fs", m = 1) + \\
    \> \>\> s(\texttt{normalized\_time}, \texttt{X}, bs = ``fs", m = 1) \\
\end{tabbing}
}

\parbox{0.48\textwidth}{
\begin{tabbing}
mm \= \texttt{pitch(logF2)} \= $\sim$ \= \texttt{gender} + \kill
    \> \texttt{pitch(logF2)} \> $\sim$ \> \texttt{gender} + \texttt{vowel} +\\
    \> \> \>s(\texttt{logF0}, by = vowel, k = 4) +  \\
    \> \> \>s(\texttt{duration}, by = vowel, k = 4) + \\
    \> \> \>s(\texttt{normalized\_time}, \texttt{speaker}, bs = ``fs", m=1) + \\
    \> \> \>s(\texttt{utterance position}, by=is.\textschwa, k = 4) + \\
    \> \> \>s(\texttt{normalized time}, \texttt{vowel sequence}, , bs = ``fs", m = 1) + \\
    \> \> \>s(\texttt{normalized time}, \texttt{X}, bs = ``fs", m = 1) \\
\end{tabbing}
}

The interaction of \texttt{normalized time} and \texttt{logF0} turned out not to be significant for F2. We therefore excluded this tensor production interaction in the GAM for F2. In the model for F1, the smooth for utterance position varies across the four vowels. For F2, however, the effect of utterance position turned out to be only modulated by whether the vowel is a schwa.

\section{Results} \label{sec: results}
All model terms were well supported. Summary tables for all the models are available in the Appendix~\ref{appendix: control variables}. In what follows, we make extensive use of visualization to clarify how the response variables are modulated by the predictors.



\subsection{Core predictors}\label{section 3.1}
In what follows, we discuss the effects of our core predictors: \texttt{place of articulation$\times$vowel}, \texttt{word}, and \texttt{word sense}.

\begin{figure}[H]
    \centering
    \includegraphics[width=1\linewidth]{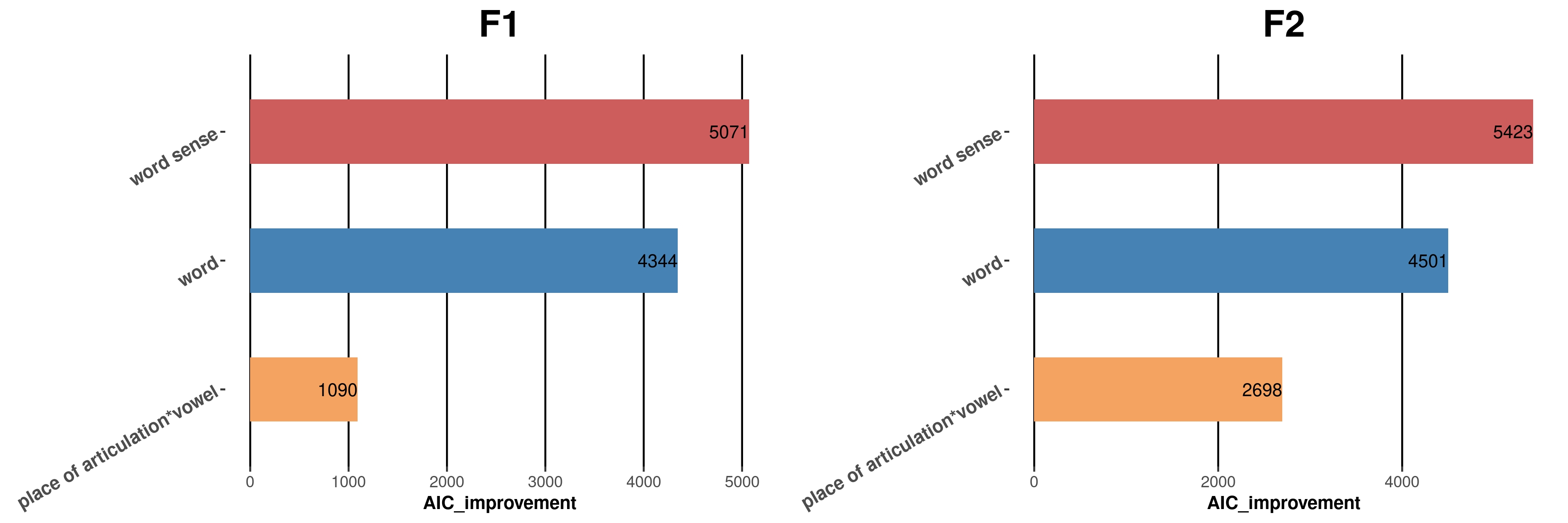}
    \caption{Improvement in model fit (using AIC), for F1 (left) and F2 (right), when adding either \texttt{place of articulation $\times$ vowel},  \texttt{word}, or \texttt{word sense} as  the core predictor to the baseline model.}
    \label{fig:AIC}
\end{figure}

\subsubsection{Place of articulation $\times$ vowel}

\begin{figure}[H]
    \centering
    \includegraphics[width=1\linewidth]{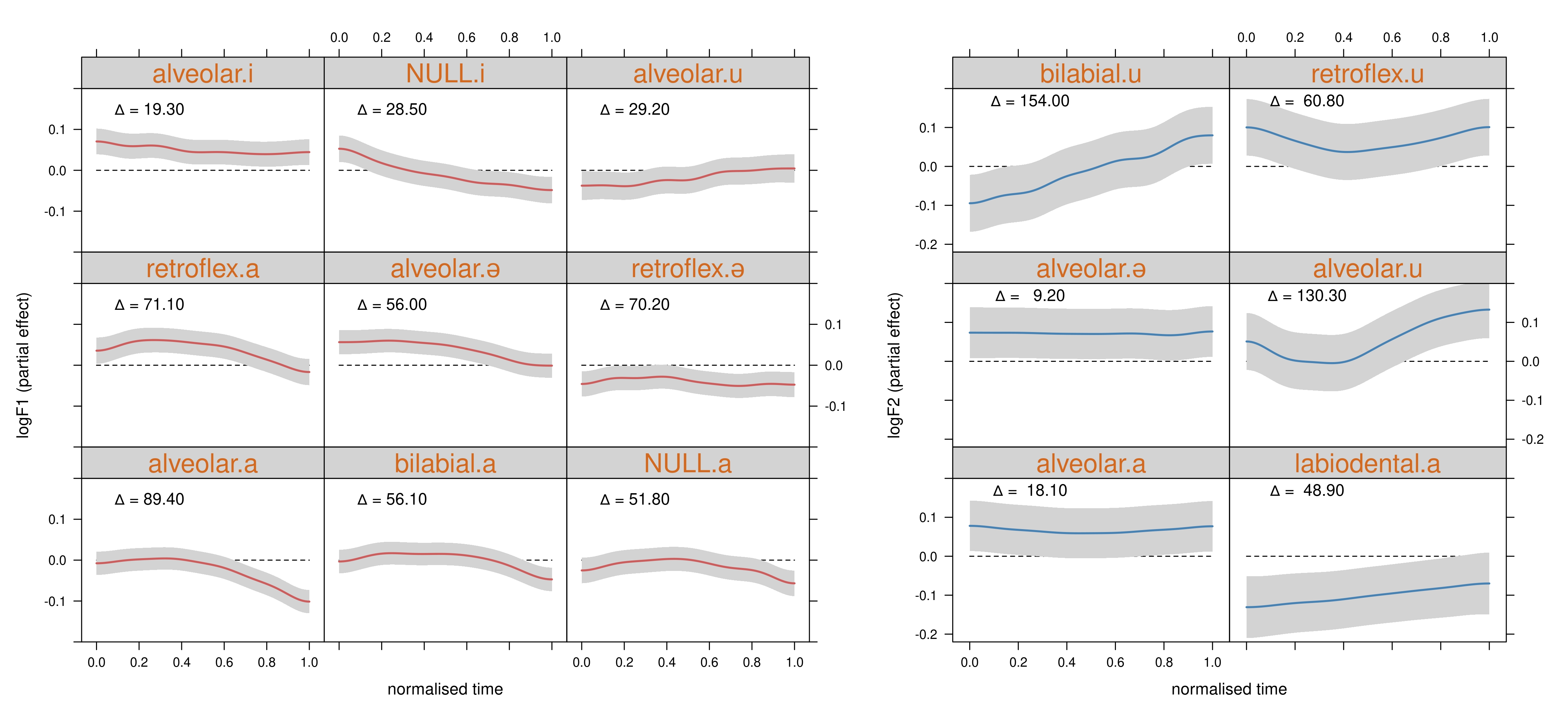}
    \caption{Partial effects for the three-way interaction of  \texttt{place of articulation $\times$ vowel $\times$ normalized time}, for F1 (left panel) and F2 (right panel). Only partial effects that are statistically significant are shown here. Panels are labelled for the combinations of place of articulation and vowel. Panels also provide the difference in Hz between the maximal and minimal values of the smooths, for female speakers.}
    \label{fig:poa}
\end{figure}

Unsurprisingly, the three way interaction of \texttt{place of articulation $\times$ vowel} by normalized time improved model fit for F1 and F2 (see the orange bars in Figure~\ref{fig:AIC}, $\Delta$ AIC: 1090 for F1 and $\Delta$ AIC: 2698 for F2, respectively). The reduction in AIC for F2 is roughly twice that for F1.  From Figure~\ref{fig:poa}, it is clear that \texttt{place of articulation} of the preceding consonant modulated formant trajectories in different ways. For /a/,  a partial effect was well supported for F1 when there was no preceding consonant  (\texttt{NULL}). When a consonant was present, partial effects were present for bilabials, alveolars and retroflexes, but not for labiodental consonants, possibly due to there being with five tokens with labiodentals (see Table~\ref{tab:poa_distribution}). 

The partial effects of F1 for /a/, shown in the bottom row of the left panel of Figure~\ref{fig:poa}, are relatively flat at the beginning and then show a downward curvature that is most pronounced for tokens with an alveolar consonant, which is in line with in British English \citep{ladgfoged2001course}. In the presence of a preceding retroflex consonant, the F1 contour is similar to that for a preceding alveolar consonant, but with a higher intercept. For the /\textschwa/, the only contours that were supported were those for words with a preceding retroflex consonant (a nearly flat but low contour) and for words with a preceding alveolar consonant (an initially flat but then  downward sloping contour). For /i/, a partial effect in the form of a downward sloping curve was supported for the condition in which there was no consonant. When an alveolar consonant preceded the /i/, the partial effect was basically level, albeit with a slightly elevated intercept, which is also seen in \cite{hillenbrand2001effects} for lab speech.   For all other consonants, partial effects were not significant.   Finally, for /u/, the only partial effect that reached support was for the tokens in which an alveolar consonant preceded the vowel. The contour is nearly flat, but a slight upward trend is present.  

For F2, fewer contours specific for the \texttt{place of articulation} of the preceding consonant reached significance. For words with /a/ that have a preceding labiodental consonant, a slight linear rise is supported, which is surprising in the light of the small number of tokens (5, see Table~\ref{tab:poa_distribution}). When an alveolar consonant precedes the /a/, the contour is level but has a higher intercept than the preceding consonant is labiodental, similar as in \cite{carden2016vowel} . A similar observation holds for the /\textschwa/ preceded by an alveolar. For the /u/, rising contours are present for words with preceding bilabial and alveolar consonants, a pattern that diverges somewhat from the English data reported in \cite{kewley1982measurement}.

\subsubsection{Word}

We fitted a GAM that included a non-linear random effect of \texttt{word} by normalised time. The AIC scores (blue bars) in Figure~\ref{fig:AIC} indicate that, for both F1 and F2, including \texttt{word} as predictor substantially improves model fit ($\Delta$ AIC = 4344 for F1 and $\Delta$ AIC = 4501 for F2 compared to the baseline models and $\Delta$ AIC = 3254 for F1 and $\Delta$ AIC = 1803 for F2 compared to \texttt{place of articulation * vowel} models).  

For /a/, 27 word types are attested in our dataset. Differences in formant trajectories and/or intercepts are clearly visible for almost all words containing this vowel (panels 1--25). For 嗎 (\textit{ma0}, panel 6, question marker) and 那 (\textit{na4}, panel 25, `that'), the F1 and F2 trajectories have significant curves and intercepts. Compared to 嗎, 那 has a lower F1 trajectory and lower F2 trajectory.  For 怕 (\textit{pa4}, panel 13, `afraid') and  殺 (\textit{sha1}, panel 19, `kill'), a reversed u-shaped curve was observed for F1, which dovetail well with the curves reported in the left panel of Figure~\ref{fig:poa} for retroflex consonants preceding /a/ and bilabial consonants preceding /a/. In contrast, 嗎 (panel 6) and 媽 (panel 10, `mother') share both the initial consonant /m/ and the vowel /a/. Nevertheless, formant trajectories differ both with respect to the intercept and with respect to shape.

\begin{CJK}{UTF8}{bsmi} 
For words with /\textschwa/ (panels 26--40), 呢 (\textit{ne0}, panel 28, sentence final particle) and 這 (\textit{zhe4}, panel 38, `this'), the word effects on the F1 trajectory's height and shape are significant. 呢 has a higher F1 trajectory than 這.  喝 (\textit{he1}, panel 19, `drink') is the only word with /\textschwa/ that has a distinctive F2 trajectory with a rising pattern. 

Panel 41--59 display the formant trajectories for words with /i/. A higher onset of F1 is visible in 裡 (\textit{li3}, panel 54, `inside') while a lowered onset of F1 is observed in 雞 (\textit{ji1}, panel 58, `chicken'). 

For words containing /u/, 煮 (\textit{zhu3}, panel 67, `cook with water') with a retroflex consonant as onset, a slight initial downward curvature is present for F2, which is consistent with the contour of \texttt{retroflex.u} in Figure~\ref{fig:poa}. A more pronounced downward trend is visible in 租 (\textit{zu1}, panel 68, `rent').

\begin{figure}[htp]
    \centering
    \includegraphics[width=1\linewidth]{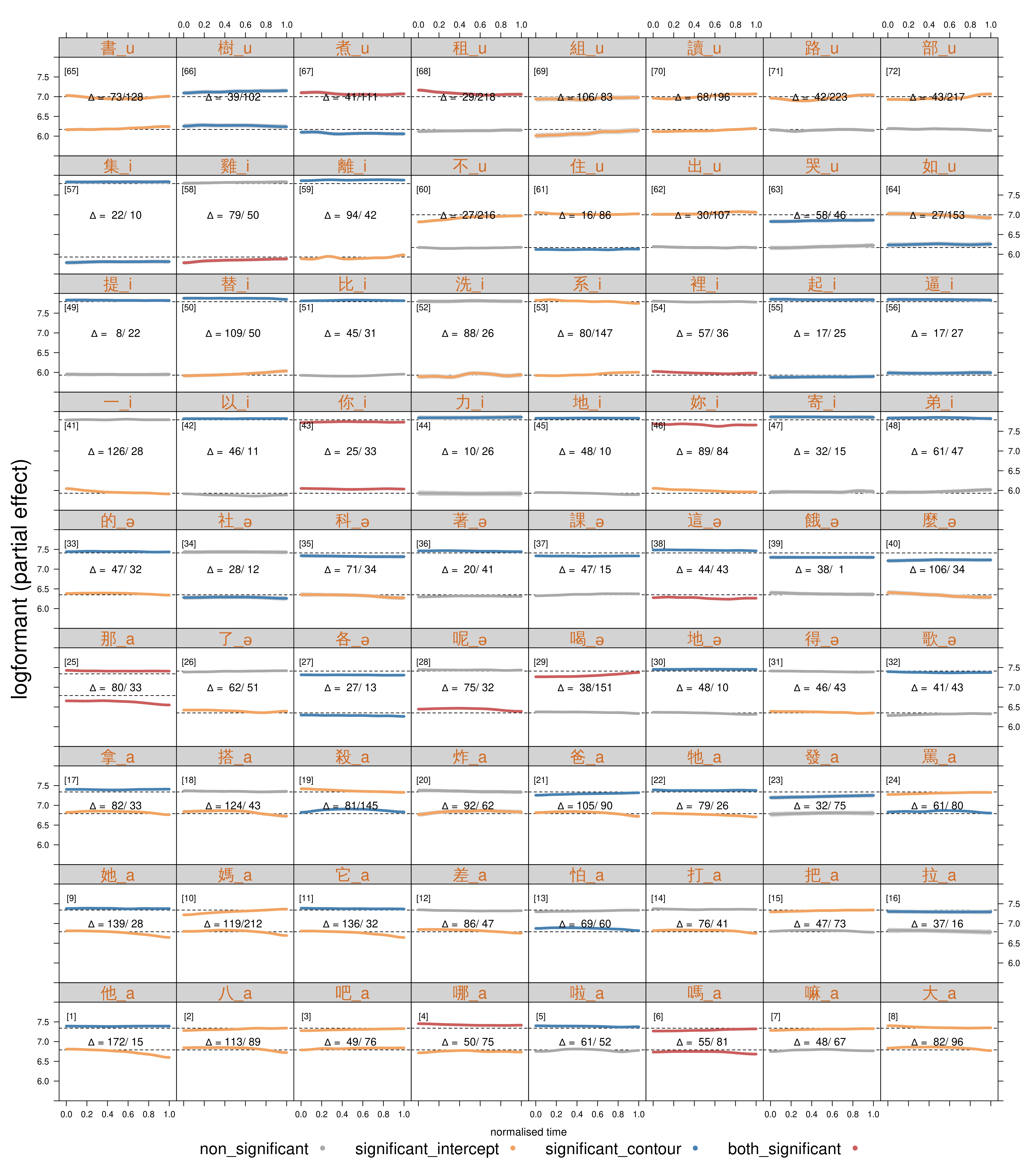}
    \caption{Partial effects of \texttt{word} in interaction with normalised time. Within each panel, the two dashed lines represent the intercepts for F1 and F2, respectively, for that vowel. Panel titles specify the words and their corresponding vowels. Statistically significant differences in intercepts are highlighted in orange, and statistically significant differences in contour shape are highlighted in blue.  Instances for which  both the intercept and contour differences are statistically significant are highlighted in red. We set $\alpha$ to 0.001 when determining the words for which a word-specific formant trajectory is well-supported}
    \label{fig:word}
\end{figure}

Figure~\ref{fig:word} presents the F1 and F2 trajectory for several sets of Mandarin heterographic homophones. For /a/, the third person pronouns,  他, 她, 它 and 牠 (\textit{ta1}, panel 1, 9, 11 and 22, `he', `she', `(inanimate) it', and `(animate) it'), have a downward trend at the end of F1. For 牠 `(animate) it', the lowering effect is slightly more moderate. The three particles containing /\textschwa/ with the same pronunciation (\textit{de0}),  地 (panel 30),  得 (panel 31) and 的 (panel 33), have different formant trajectories. F1 trajectories for 得 and 的  are higher than the mean F1 value for /\textschwa/ (indicated by the lower horizontal dashed line), and 的 and  地 have significant movement for F2, whereas 得 has a stationary F2. The word 你 (\textit{ni3}, `you (male)', panel 43) has a higher F1 trajectory and lower F2 trajectory compared to 妳 （`you (female)', panel 46). This shows that heterographic Mandarin words may have different vowel formant trajectories. 

\end{CJK}

\subsubsection{Word sense}

The results of the preceding section supported our hypothesis that the effect of \texttt{word} is stronger than the effect of the place of articulation of the preceding consonant. We now consider whether \texttt{word sense} is a better predictor of formant trajectories than \texttt{word}. To do so, we replace the by-\texttt{word} factor smooths for normalized time with by-\texttt{word sense} factor smooths. 

Figure~\ref{fig:AIC} (presented above in section~\ref{section 3.1} presents the improvement in AIC when \texttt{word sense} (red bars) is added to the baseline model. The AIC improves both for F1 (by 5071 AIC units) and F2 (by  5423 AIC units), with a smaller improvement for F1.   The improvement of the \texttt{word sense} GAM compared to \texttt{word} GAM ($\Delta$ AIC = 727 for F1 and $\Delta$ AIC = 922 for F2) supports the possibility that indeed word meaning co-determines formant trajectories. 

\begin{figure}[htp]
    \centering
    \includegraphics[width=1\linewidth]{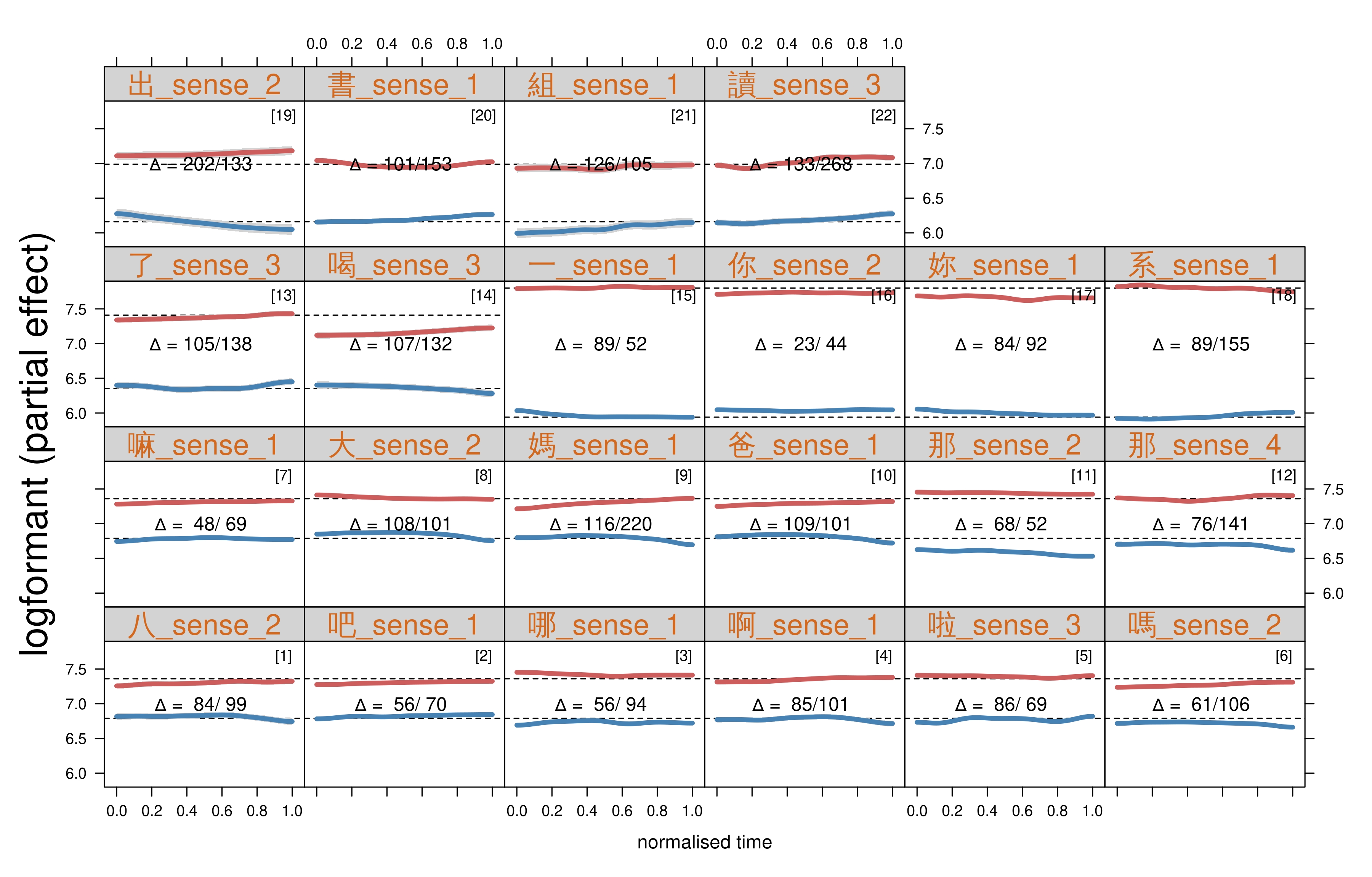}
    \caption{Partial effect of \texttt{word sense} in interaction with normalized time. Contours are shown for those senses which have statistically significant partial effect smooths for both F1 and F2.}
    \label{fig:sense}
\end{figure}

\begin{CJK}{UTF8}{bsmi} 
Figure~\ref{fig:sense} displays the partial effect of \texttt{word sense} for all cases where clear sense-specific contours are present for both F1 and F2. For 那 (\textit{na4}), the formant trajectories of the two senses are displayed in panels 11--12. With the meaning of `that' (panel 11), as in 那種領域 (\textit{na4-zhong3-lin3yu4}, `that field'), the F1 trajectory is lower, while the F2 trajectory is higher, compared to the formant trajectories of the second sense, which is `to mark the change in tone in speech' as in 那這隻狗真的很奇怪喔 (\textit{na4-zhe3-zhi1-gou3-zhen4de0-hen3-qi2guai4-wo0}, `Then this dog is really strange').  For 喝 (\textit{he1}) with the meaning `drink (alcohol)' (panel 14), the F2 trajectory is significantly different from the mean F2 trajectory of words with /\textschwa/ and from the F2 trajectories in the other meanings of 喝 (such as `drink tea'). The two second person pronouns 你 (panel 16) and 妳 (panel 17) have distinct trajectories compared to other words with /i/.  The word 讀 (\textit{du2)} is attested with three meanings in our dataset. In panel 22, with the meaning of `read magazines', its F1 trajectory has an upward trend and F2 first declines then levels off. For the other meanings of 讀, differences in contours are absent for both F1 and F2. 
\end{CJK}

\begin{figure}
    \centering
    \includegraphics[width=1\linewidth]{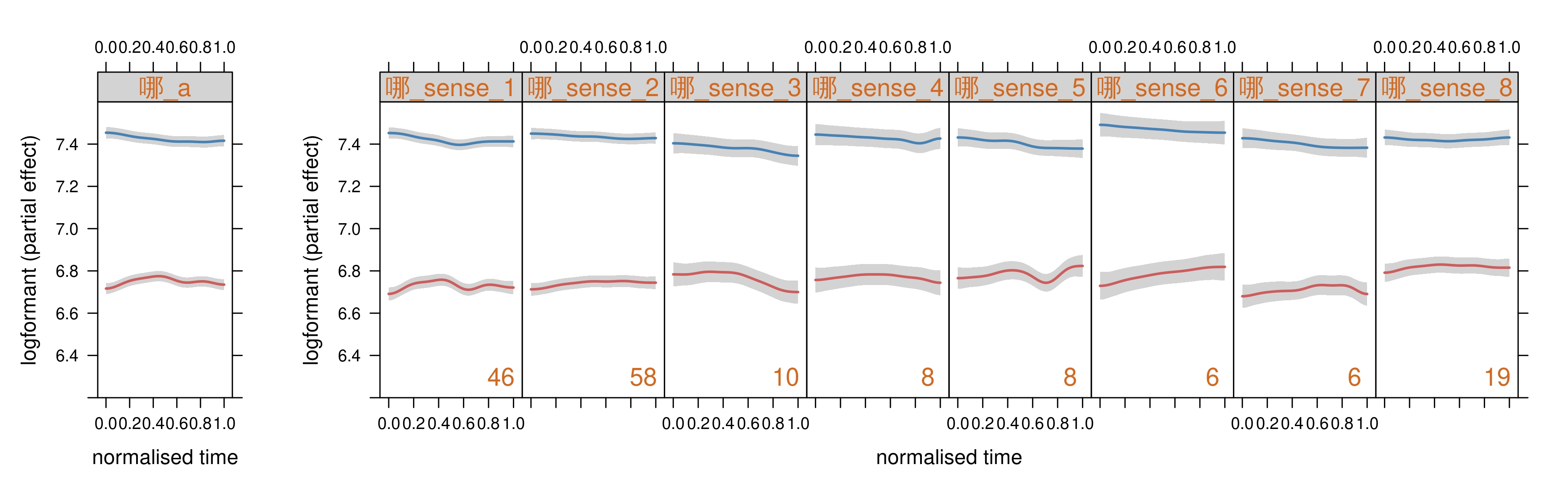}
    \caption{Factor smooths for \texttt{word} (left panel) and \texttt{word senses} (right panel) for `哪'.  (Factor smooths include differences in the intercept.)  The numbers in the lower right of each panel are the token identifiers in our dataset.  The smooth from the word-based GAM is the average of the smooths from the sense-based GAM.}
    \label{fig:na}
\end{figure}

The predictor \texttt{word} has 87 levels while \texttt{word sense} has 188 levels. As a consequence, the greater variable importance of \texttt{word sense} may be an artifact of the higher number of levels of word sense. To investigate whether this is the case, we studied the partial effect for the word `哪', which has the biggest number of word senses in the current study, both in the \texttt{word} GAM and from the \texttt{word sense} GAM. These are shown in Figure~\ref{fig:na}.  As the number of tokens for senses 3, 4, 5, 6 and 7 are modest, confidence intervals are wider compared to those of senses 1 and 2. The trajectories of the word GAM (left panel) are roughly the average of the different  sense-specific trajectories (right panel).  The dipping pattern for the word-based F1 can be traced back to the dipping curves of senses 1 and 5. Likewise, the height of the F2 in the word GAM is a compromise between the low intercepts of senses 3 and 5, and the much higher intercepts of the other senses. This pattern of results suggests that the sense analysis and the word analysis are not contradicting each other, and that the word sense analysis offers a more fine-grained window on the realization of the formant trajectories.

\subsection{Cross validation}
In order to ascertain whether the present results are robust, we carried out a cross-validation study. We held out 10\% of the current data as test data, and used the remaining 90\% as training data. Every \texttt{word type} was represented in both the training and test data. The number of tokens per word type in the test data was proportional to that in the training data. We did so 10 times, for each of our statistical models: the baseline model, the \texttt{place of articulation $\times$ vowel} model, \texttt{word} model and the \texttt{word sense} model. To quantify model accuracy, we obtained the models' predictions based on the training data for the formant values of the held-out test data, and calculated the sum of squared errors (SSE). 

\begin{figure}
    \centering
    \includegraphics[width=1\linewidth]{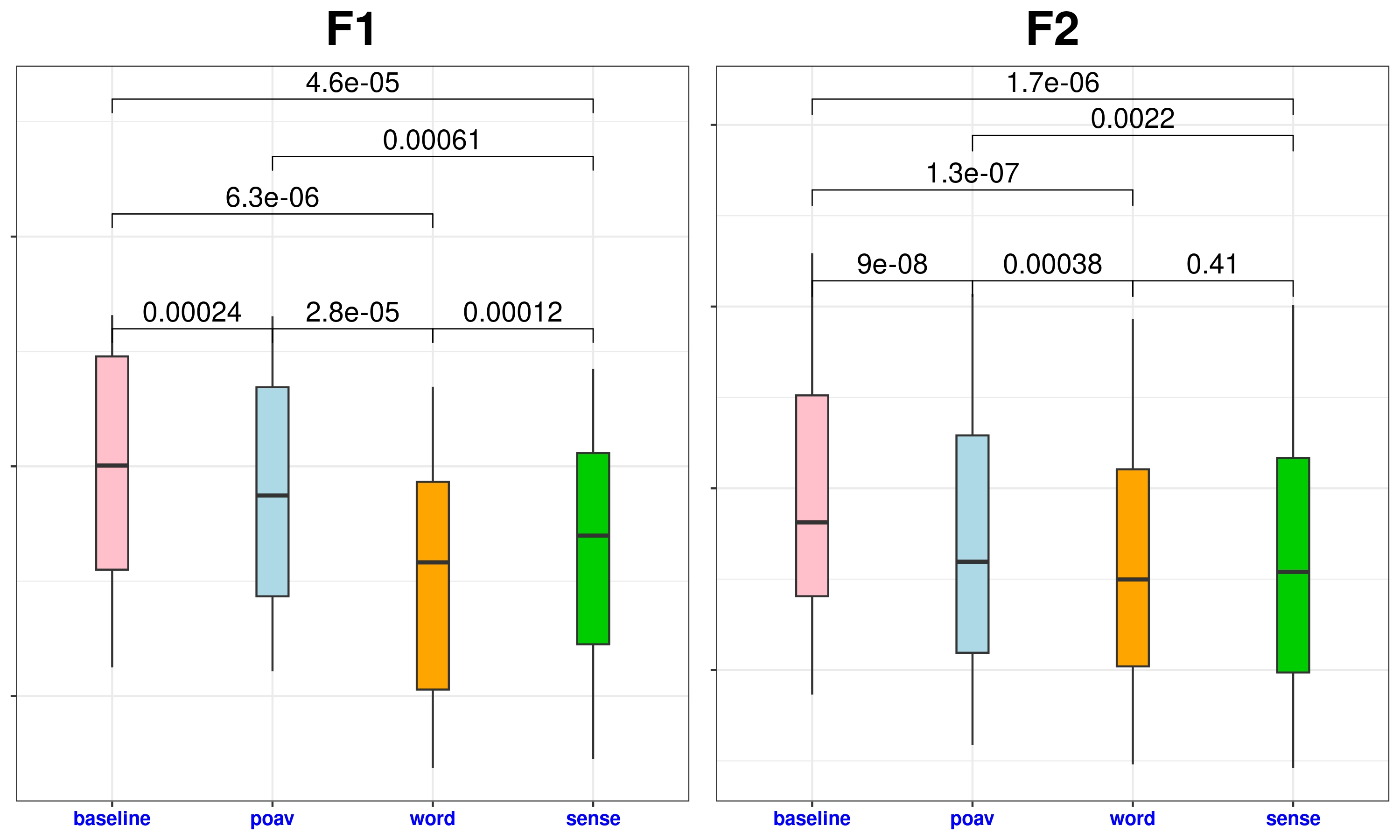}
    \caption{SSE (sum of squared errors) in 10 cross validation runs for the baseline GAM model (pink) and GAM models with three different:   \texttt{place of articulation $\times$ vowel} (poav, blue), \texttt{word} (orange), and \texttt{word sense} (green). The p-values between pairs of models are derived from paired t-test. }
    \label{fig:cross validation}
\end{figure}

As shown in Figure~\ref{fig:cross validation}, the SSE of the \texttt{word} GAM model was lower than that of the baseline model, for both F1  ($t_{(9)} = -9.3421, p < 0.0001$) and F2 ($t_{(9)} = -14.812, p < 0.0001$). The SSE for F2 of the \texttt{word sense} GAM model was indistinguishable from that of the \texttt{word} model ($t_{(9)} = 0.8708, p = 0.4065$). 
However, the SSE for F1 of the \texttt{word sense} GAM was lower than the SSE of the \texttt{word} GAM ($t_{(9)} = 6.4484, p = 0.0001$). 
The \texttt{word sense} model always improves on the \texttt{baseline} model ($t_{(9)} = -7.291, p < 0.0001$ for F1 and $t_{(9)} = -10.938, p < 0.0001$ for F2) and the SSE of a GAM model with the \texttt{place of articulation $\times$ vowel} interaction ($t_{(9)} = -5.1393, p = 0.0006$ for F1 and $t_{(9)} = - 4.2251, p = 0.0022$ for F2). 

There are several possible reasons for why the performance of the \texttt{word sense} model is not better than that of \texttt{word} model in the case of F1. First, monosyllabic words are highly polysemous. Some words in our dataset are function words, which challenge sense tagging. Second, from a statistical perspective, the reduced precision of the \texttt{word sense} model for held out data is unsurprising:  there are many more  \texttt{word senses} than words, and each of these word senses is supported by fewer word tokens than the \texttt{words} (characters).  Especially for word senses with relatively few tokens, training on even fewer tokens is highly unlikely to result in higher prediction accuracy \citep[see also][for detailed discussion of a similar finding for F0]{Chuang_Bell_Tseng_Baayen_2026}. 

\subsection{Discriminative lexicon model}

The results of the GAM modelling supports our hypothesis that vowel formant trajectories are co-determined by words' meanings.  To provide stronger support for the effect of word senses on formant contours, we tested whether the Discriminative Lexicon Model (DLM) \citep{baayen2019discriminative,chuang2021discriminative,Heitmeier_Chuang_Baayen_2026} can predict vowel formant trajectories from words' meanings, approximated with contextualized embeddings.

The DLM has been used successfully to capture the alignment between distributional semantics and different types of fine-grained phonetic variation across different languages, such as the spoken word durations of homophones in English \citep{Gahl_Baayen_2024} and tonal realization  in Mandarin words \citep{Chuang_Bell_Tseng_Baayen_2026,lu2025realization,JIN2026101495}.  The DLM represents words' forms and meanings with high-dimensional numeric vectors, and defines mappings that predict meaning vectors from form vectors (comprehension), and vice versa (production).  The DLM modelling framework predicts that it should be possible to predict the formant trajectories of word tokens from their meanings in context. If this prediction turns out to be correct, this provides further support for the effect of word being a semantic effect. 

In the current study, the meanings of words in context are represented by means of contextualised embeddings (CEs), obtained with a pre-trained unidirectional language model based on the GPT-2 architecture, developed by CKIP \footnote{ckiplab/gpt2-base-chinese, available at https://github.com/ckiplab/ckip-transformers.}).  Each token in our dataset is paired with a 768-dimensional vector representing its context-specific meaning. In a DLM model, the form vectors should have the same dimensionality as the meaning vectors. However, the empirical formant vectors have different lengths:  the formant values were measured every 2 ms, which resulted in different numbers of measurement points for tokens differing in their spoken word duration. To sidestep this problem, we used the predicted formant values from the \texttt{word} GAM.  We used this GAM model to generate formant predictions at 100 equally spaced time points ranging between 0 and 1 for every individual token.

A speaker's individual characteristics, such as vocal tract, health, and emotional state at the time of recording, likely  influenced their speech. However, the CEs are based entirely on written text, and are blind to differences between individual speakers.  To reduce the effects of individual differences, we centred the predicted formant values by token. For each token, we calculated the  mean of the 100 predicted values, for F1 and F2, respectively, and subtracted the mean value from each predicted formant value. As a consequence, with centred predicted formant values, the effects of speakers having different vocal tracts are removed, while leaving the shape of the formant contours untouched.  In what follows, we refer to these formant vectors as `gold-standard' formant vectors.

The input to the DLM mappings was a matrix with the CEs of the tokens as row vectors. The DLM was given the task of mapping this matrix onto a matrix with as row vectors the gold-standard formant vectors. Two mappings were created, one for F1 and one for F2. Given these two mappings, the model generated predicted formant vectors, one for F1 and one for F2.  

For evaluation, we carried out 10 cross-validation runs. For each run, we split our dataset into a training dataset (90\%) and a testing dataset (10\%). Word types and word senses were always present in both the training and the test data; the numbers of tokens were chosen such that counts of occurrences were proportional for training and testing.  Mappings between meanings and forms were calculated using the training dataset and evaluated on the testing dataset. 

To gauge the precision of these mappings, we calculated the euclidean distance between the gold-standard and predicted F1 and F2 vectors, respectively. If the nearest neighbour of a predicted formant vector was a token of the same word type as the target token, the predicted formant vector was assessed as correct. In addition to calculating accuracies for the two formants separately, we also calculated the accuracy based on whether, for a given word token, both F1 and F2 were predicted accurately. In order to establish a baseline for comparison, we carried out exactly the same modelling steps, but now for datasets in which the order of the CEs was randomly permuted.

\begin{figure}
    \centering
    \includegraphics[width=0.8\linewidth]{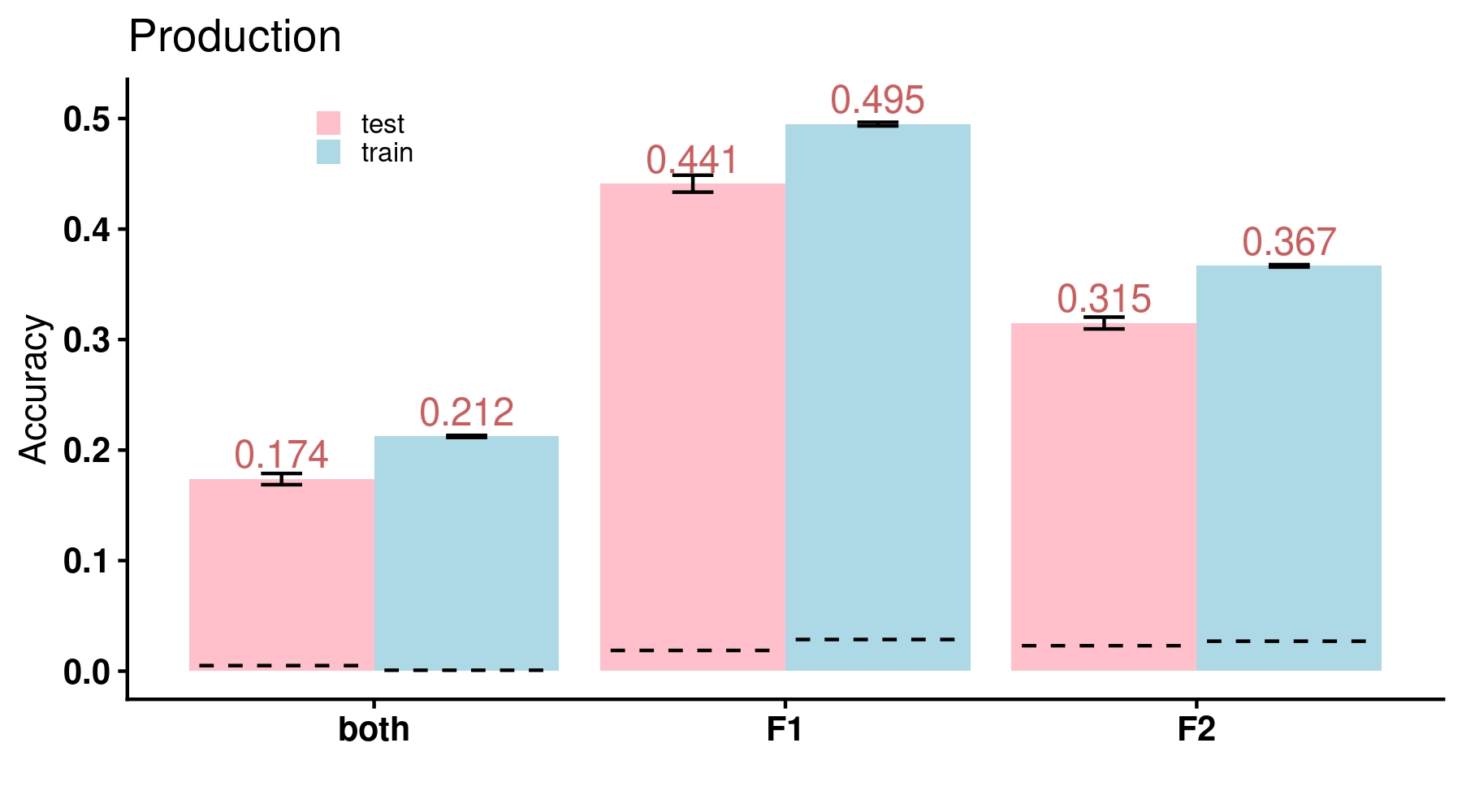}
    \caption{Mean production accuracies for training data (blue bars) and testing data (pink bars) from DLM. Mean accuracy is obtained from 10 random training and testing splits. Dashed  lines represent global permutation baselines.}
    \label{fig:production}
\end{figure}

As shown in Figure~\ref{fig:production}, the accuracy rates in training are higher than in testing. For F1, the corresponding accuracies are 49.5\% in training and 44.1\% in testing, respectively. The accuracy for F1 is a little higher than the accuracy for F2 (36.7\% in training and 31.5\% in testing). For the `both' condition (when the nearest neighbours of both F1 and F2 are required to be tokens of the same type),  the accuracy in training and testing is 21.2\% and 10.8\%, respectively. Although the accuracy for `both' is much lower than for F1 or F2 alone, the accuracy is above the permutation baseline, which is close to zero (0.08\% for training and 0.05\% for testing). \footnote{The accuracy results for original formant (non-centered) values are presented in  Appendix~\ref{appendix: non-centered DLM}. The results for non-centred formant values holds for the same as centred values presented here.} 


In summary, we have shown that the formant trajectories realized on the word tokens in our dataset can be predicted from their corresponding contextualized embeddings, with an accuracy that substantially exceeds a randomization baseline. This result fits well with our hypothesis that the word effect we observed in the GAM analyses is a semantic effect.  Apparently, meaning in context and the details of how formants are realized are calibrated at the level of individual tokens.

\begin{CJK}{UTF8}{bsmi} 

\begin{figure}
    \centering
    \includegraphics[width=1\linewidth]{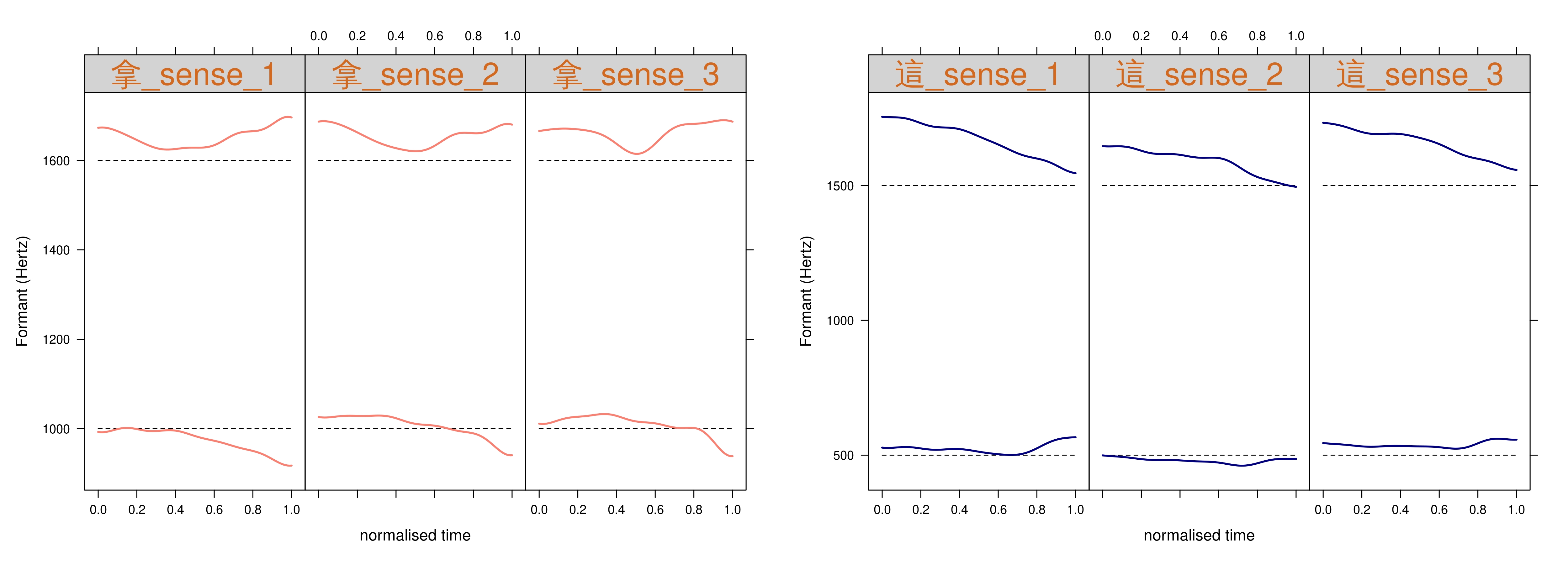}
    \caption{Predicted formant trajectories from the \texttt{word sense} GAM model for three word senses of 拿 for a female speaker (left panel) and for three word senses of 這 for a male speaker(right panel). Predictions are for median \texttt{duration} for each vowel, with \texttt{speaker} set to the first male and female speaker in the dataset. \texttt{Utterance position} is set to 1, with the \texttt{adjacent vowel} fixed at `NULL-NULL'. }
    \label{fig:visc}
\end{figure}

We have shown that word meaning co-determines the fine phonetic details of the formant trajectories of four vowels in Taiwan Mandarin. It follows that vowel formants must exhibit word-specific temporal trajectories. This is supported by  two word types that are attested with three different senses in our dataset and for which we obtained the predicted formant values from the \texttt{word sense} GAM model.  As can be seen in Figure~\ref{fig:visc},  F1 and F2 change over time in slightly different ways for the different senses. The left panel of Figure~\ref{fig:visc} presents the predicted formant trajectories for `拿' (\textit{na2}, `take') for a  female speaker (left panels), and the right panel for 這 (\textit{zhe4}, `this')  for a male speaker (lower panels). The senses of these words are listed in Table~\ref{tab:sense_translation} in the Appendix~\ref{appendix: translation}.  In the case of 拿, for sense 1, F1 shows an earlier onset of the downward trend, and for sense 3, F2 shows a more pronounced dip at 0.5 normalized time. In the case of 這, sense 1 and sense 3 have higher F2, and sense 3 has slightly higher F1.

\end{CJK}

\section{General Discussion}\label{sec: gendisc}






The present study investigated vowel inherent spectral change (VISC) for /a, i, u, \textschwa/ in  spontaneous conversations in Taiwan Mandarin. VISC has previously been documented for some European languages, such as different varieties of English, German, Dutch and Spanish \citep[e.g.,][]{nearey1986modeling,hillenbrand1995acoustic,brandt2021dynamic,williams2015beyond,elvin2016dynamic, morrison2009l1}, and also for Mandarin Chinese \citep{yuan2013spectral}. Our results complement the latter study, which investigated laboratory speech, with an investigation of spontaneous speech.  Our results show that VISC is present not only in laboratory speech,  but also is a robust feature of spontaneous spoken Mandarin. 

A novel finding of the present study is the existence of word-specific components of VISC contours that can be isolated from the effects of prosodic and sociolinguistic factors. Importantly, words' word-specific formant trajectories are predictable from their corresponding contextualized semantic embeddings with an accuracy that exceeds a permutation baseline. This finding adds a new perspective on the factors shaping VISC, complementing explanations that are grounded in the physics of speech and the realization of  phonological contrasts discussed in previous research \citep{nearey1986modeling,morrison2012theories,sims2012modelling}.

In what follows, we return to the three research questions laid out in the introduction, and 
elaborate on the theoretical and methodological implications of our results.

\begin{enumerate}
    \item Following up on an earlier research of VISC in Mandarin laboratory speech \citep{yuan2013spectral}, by analysing thousands of word tokens taken from spontaneous conversations, we observed systematic F1 and F2 trajectories changes over time.  The magnitude of VISC in conversational speech in some tokens is amplified compared to previously reported lab-speech values, which is due to the limitation of the LPC algorithm when applied to spontaneous speech. However, the spectral change across different vowel types remains a stable phenomenon. The fact that VISC is observed in conversational Mandarin speech, which has reduced clarity and overlapping co-articulatory pressures, extends the empirical scope of VISC.  This answers \textbf{research question 1, which asks whether VISC is also present in conversational Mandarin}.
    
    \item Using the generalized additive model \citep[GAM][]{Wood:2017}, we decomposed the vowel formant trajectory into prosodic factors that have been observed to be effective predictors of formant dynamics, such as duration, prosodic position in utterance and predictability in context, in combination with sociolinguistic factors such as speaker gender and speaker identity. Our \textbf{second research question asked whether VISC contours have a word-specific component, similar to the word-specific components reported for F0 in Mandarin} \citep{Chuang_Bell_Tseng_Baayen_2026,lu2025realization,JIN2026101495}. The GAM results clearly support the possibility that VISC has a word-specific component, even when taking the contribution of prosody into consideration (details see Appendix~\ref{appendix: control variables} and Appendix~\ref{appendix: bigram}).     
    We hasten to note here that we do not claim that the VISC is only determined by word meaning.  Word meaning is a novel co-determinant of VISC, extending recent findings for F0 to F1 and F2.

    \item \textbf{Our third research question asks whether word-specific VISC components are to some extent predictable from words' contextualized embeddings}, as predicted by the Discriminative Lexicon Model \citep{Heitmeier_Chuang_Baayen_2026}.  We used a simple linear mapping from GPT2-based contextualized embeddings to time-normalized formant values predicted by GAMs.  Although linear mapping may turn out to be too simple, given the current state of knowledge, they allow us to isolate any systematic covariation between phonetic trajectory shape and semantic space without introducing non-linearities that might overfit and wrap themselves around spurious patterns. We show that a GAM that includes word type as a predictor, when trained on 90\% of the data, provides predictions for the formant trajectories of the held-out data that have a precision that far exceeds a permutation baseline.  We do not take this result to imply that language users are aware of how meaning affects the realization of formants, or that they would be able to use this information in meta-linguistic tasks.  What the existence of word-specific and meaning-specific VISCs does indicate is that, apparently, listeners absorb the statistical relations between meaning and form from their interactions with their interlocutors, and as speakers learn to fine-tune their speech to the specific meanings they are communicating. 
\end{enumerate}

A series of studies have proposed explanations for the existence of VISC.  According to  \cite{lindblom1963spectrographic,lindblom1967role}, speakers may fail to reach a static articulatory target due to temporal constraints: the jaw, tongue, and lips must rapidly transition between neighbouring segments in context.  However, subsequent studies demonstrated that substantial spectral movement is persistent even in citation-form speech, for carefully articulated vowels \citep{hillenbrand1999identification,hillenbrand2001effects}. In current study, the effect of the consonants around the vowel has been controlled for statistically in the  GAMs.  Word-specific formant trajectories remain clearly visible, indicating that VISC cannot be explained solely by co-articulatory undershoot.

Within articulatory phonology, VISC emerges naturally from the biomechanical properties of speech production, including tongue inertia and the temporal coordination of articulatory gestures \citep{Browman:Goldstein:1986,Tiede:2011}. 
%
%
%
This suggests that possibly the observed word-specific formant trajectories observed in current study would be the acoustic reflexes of stable, word-level gestural patterns, shaped by motor constraints and accumulated experience. However, the current study provides evidence that the formant trajectories of individual word tokens can be predicted with an accuracy substantially above a randomization baseline from words' contextualized embeddings, using a linear transformation from the embedding space to the time-normalized formant space. This finding that words' meanings in context also shape their phonetic realization introduces a new factor beyond physiological and articulatory factors and motor practise: lexical semantics. 
Our result is surprising in the light of the substantial between-speaker variability in articulation 
on the one hand, and on the other hand, the lack of precision in the contextualized embeddings, which are based on a large language model that is blind to the specific meanings, emotions, and pragmatic targets that individual speakers in our corpus may have had in mind.


We acknowledge that contextualized embeddings derived from large language models capture a broad range of linguistic information beyond pure semantics. Nevertheless, the primary and most robust signal encoded in these embeddings remains semantics. For instance, a linear discriminant analysis \citep{fisher1936use,Venables:Ripley:2002} using leave-one-out cross-validation, given word embeddings, predicts word type (87 different types) correctly for 95\% of the tokens, Part of Speech (15 classes) with 96\% accuracy, whether a word is sentence initial in 81.8\% of the cases, and whether it is sentence final in 73.0\% of the tokens \citep[for a study predicting aspects of prominence from embeddings, see][]{talman2019predicting}. 

The novel finding that words' embeddings also modulate formant trajectory changes over time, does not imply that other well-established effects of prosodic factors or articulatory constraints are not shaping formant dynamics. Instead, we argue that distributional semantics operates in addition to these constraints: for a given prosodic context and a given duration, vowel formant trajectories are predictable to some extent also from their meaning in context. 

The present approach to articulation differs from standard theories according to  which segments define targets for the individual articulators with smoothed between-target transitions.  The alternative that is explored in the present study and related work proposes that speakers have articulatory targets for shape and duration, without requiring segment-based processes \citep[cf.][for a critique of phonological segments]{Port:Leary:2005}.  Whether this approach is cognitively valid (the independence of shape and duration estimation may be too simplistic) and can be made to work for articulation in all its complexity is an open question. 

It has been noted that spectral movement may enhance vowel identification by providing listeners with more information \citep{morrison2007testing,strange2012dynamic}.  Our findings add to this the possibility that VISC is not merely a passive consequence of articulatory constraints that work out beneficially for the listener, but that VISC in addition is modulated by semantics, and contributes guiding the listener to the critical region in semantic space where a word's meaning lives.

%

In current study, we analysed formant trajectories in normalized time. It might be argued that analyses in normalized time cannot capture articulatory details such as the speed and acceleration of the articulators.  Indeed, time normalized VISC trajectories do not reflect speed and acceleration, but if the characterization of their shape in normalized time is correct, then when time-normalized trajectories are rescaled back to time in ms, then speed and acceleration can be accurately estimated.   Interestingly, spoken word duration can also be predicted in part from word embeddings, as shown by \citet{Gahl_Baayen_2024} and \citet{jin2026usingembeddingspredictspoken}. We leave it to further research to clarify whether combining models for VISC in normalized time and for spoken duration in ms will succeed in approximating well the VISCs on the ms timescale. 


\clearpage
\appendix
\section*{Appendix}
\setcounter{figure}{0}
\setcounter{table}{0}

\section{Control variables} \label{appendix: control variables}
The first four panels in the first row of Figure~\ref{fig: time and logF0} plot the partial effect of log-transformed fundamental frequency (F0) on the trajectory of log-transformed F1. For /a/ and /i/,  an increase of F0 goes hand in hand with an increase in F1 while for /\textschwa/ and /u/, F1 first goes down and then increases, resulting in  a U-curve. The last panel in the first row displays the partial effect of the interaction between log-transformed F0 and normalised time.  Darker red colours denote higher formant values. When F0 increases, F1 increases as well. Apparently, this increase is modulated slightly over time, such that at vowel offset, F1 is lowest for intermediate values of F0.  

The first four panels in the second row  visualize how formant trajectories are modulated by  duration.  For high vowel such as /i/ and /u/, with longer vowel duration, F1 decreases while for mid vowel /\textschwa/ and low vowel /a/, their F1 increases. The last panel in the second and the first three panels in the third row illustrate the partial effect of utterance position on F1. For /a/, a fall is initiated halfway through the vowel. For \textschwa/ and /i/, F1 decreases and then rises.  For /u/, F1 first decreases and then levels off.  

The two rightmost panels in the third row and the first panel in the last row of Figure~\ref{fig: time and logF0} show how F2 is influenced by F0. A linear downward trend is visible for /u/. For /a/ and /\textschwa/ a reversed u-shaped pattern is present. The following three panels describe the modulation of F2 by duration. For /a/ and /u/, longer vowel duration results in decrease of F2. The last panel of Figure~\ref{fig: time and logF0} shows that when positioned closer to the middle of the utterance, F2 of /\textschwa/ increases and then levels off when the vowel is approaching to the end of utterance.

\renewcommand{\thefigure}{A\arabic{figure}}
\begin{sidewaysfigure}
    \centering
    \includegraphics[width=1\linewidth]{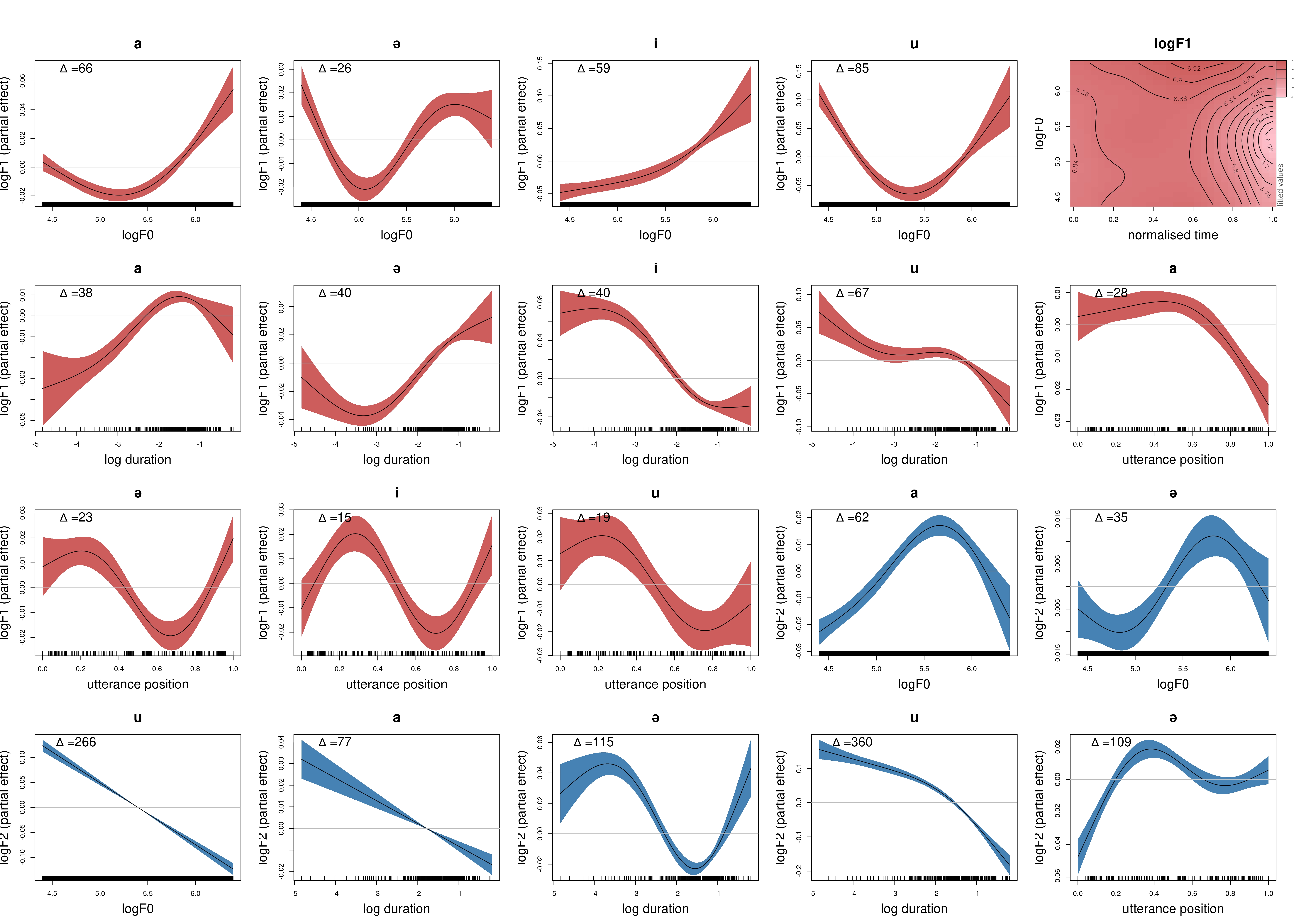}
    \caption{Significant partial effects from the baseline GAM models for F1 and F2. The red panels in the first, second and the third row present the partial effects of the control variables for F1 and the blue panels in the third and last row present  partial effects of the control variables for F2. All partial effects are supported by p-values $< 0.0001$. The number in the upper-left corner of each panel indicates the difference in Hz between the lowest and highest values, back-transforming from predicted log formants.}
    \label{fig: time and logF0}
\end{sidewaysfigure}

\section{Word senses in Figure~\ref{fig:visc}}\label{appendix: translation}
\begin{CJK}{UTF8}{bsmi}
In Figure~\ref{fig:visc}, we selected two Mandarin monosyllabic words (`拿' and `這') that are attested with three senses. Complementing Figure~\ref{fig:visc}, the following table lists the word senses and their definitions in
Mandarin and English.

\renewcommand{\thetable}{A\arabic{table}}
\vspace*{2\baselineskip}
\nopagebreak
\footnotesize{
\begin{longtable}{p{3cm}p{6cm}p{6cm}}
\hline 
\textbf{Character\_Sense} & \textbf{Sense Meaning} & \textbf{Translation}\\\hline
          
拿\_sense\_1 &  引介事件所憑藉的方法或工具 & the method or means by which an event is introduced\\
          
拿\_sense\_2 &  向特定對象取得主事者擁有或應該擁有的金錢或物品 & obtain money or goods that the agent possesses or ought to possess from a specific target \\

拿\_sense\_3 &  用手取物或持物 & take or hold an object with the hand \\

這\_sense\_1 &  根據說話者說話時所處的時空而言，指稱比較近的後述對象 & referring to a subsequent referent that is relatively close, with respect to the speaker’s spatiotemporal context at the time of speaking \\

這\_sense\_2 &  代指上文中已提過的對象 & refer back to an entity previously mentioned in the discourse \\

這\_sense\_3 &  表言談中用於標記語氣的停頓 & a pause used in speech to mark tone or discourse stance \\

 \hline
    \caption{Word senses referenced in Figure~\ref{fig:visc}.}
    \label{tab:sense_translation}
\end{longtable}
}
\end{CJK}

\section{DLM: accuracy rate for original formant values (non-centered)}\label{appendix: non-centered DLM}

For the original GAM-predicted formant vectors, the accuracy exhibits a similar trend to that of the centered formant vectors. Training accuracy consistently exceeds testing accuracy across all conditions. For both training and testing sets, F1 yields higher higher accuracy than F2. In the `both' condition, although the accuracy is lower than that for predicting F1 or F2 alone, it still significantly outperforms the permutation baseline.  

A comparison of results from the two types of formant vectors (original vs centred) showed that prediction is consistently more accurate when the formant trajectories are centred. By centering the formant trajectories, the differences between individual speakers, which are invisible to the GPT model generating the CEs, are filtered out, and the general patterns in the data emerge more clearly.

\begin{figure}[H]
    \centering
    \includegraphics[width=0.8\linewidth]{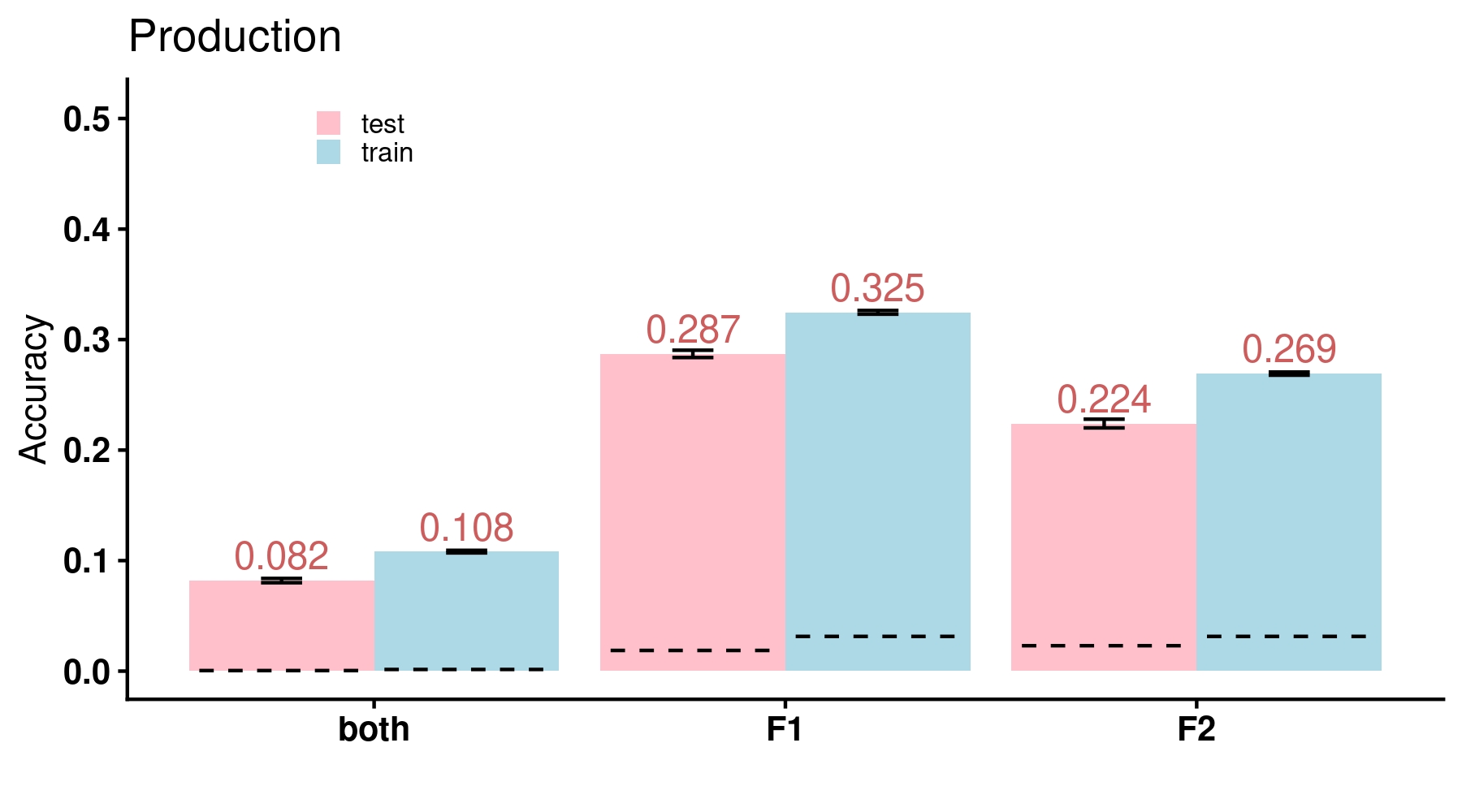}
    \caption{Mean production accuracies for training data (blue bars) and testing data (pink bars) from DLM. Mean accuracy is obtained from 10 random training and testing splits. Dashed lines represent global permutation baselines.}
    \label{fig: LDL_noncentered}
\end{figure}

\section{Bigram probability}\label{Bigram}\label{appendix: bigram}
In previous studies that examines the meaning of the word on the phonetic realizations of Mandarin disyllabic words \citep{Chuang_Bell_Tseng_Baayen_2026,lu2024form}, \textbf{bigram\_probability} was included as a control variable. Some studies found that higher
predictability is associated with shorter word duration and a greater degree of spectral reduction \citep{gahl2012reduce,bell2003effects}. For tonal realizations in Mandarin, \cite{Chuang_Bell_Tseng_Baayen_2026} found that the F0 excursion of disyllabic Mandarin words were reduced with higher values of bigram probabilities. 

Therefore we tried to add bigram-probability into GAM. Following \cite{Chuang_Bell_Tseng_Baayen_2026}, we calculated the \texttt{bigram probabilities} of target tokens in two ways: \texttt{bigram\_previous}, based on the
preceding word, and \texttt{bigram\_following}, based on the following word. These two variables are defined, respectively, as follows:

\[
P(w_n \mid w_{n-1}) = \frac{\text{Freq}(w_{n-1}, w_n)}{\text{Freq}(w_{n-1})}
\]

\[
P(w_n \mid w_{n+1}) = \frac{\text{Freq}(w_n, w_{n+1})}{\text{Freq}(w_{n+1})}
\]

where $P(w_{n} \mid w_{n-1})$ is the probability of a word occurring given the previous word, $P(w_{n} \mid w_{n+1})$ is the probability of a word occurring given the following word, and $Freq$ denotes word frequency in the corpus of Taiwan Mandarin.

As the distribution of bigram probabilities have a long right tail, a transformation of the bigram probabilities is therefore required. We selected the natural log transformation to reduce the skewness of bigram probabilities. We then added two additional smooth terms and tensor product interaction smooths, for \texttt{log\_bigram\_previous} and \texttt{log\_bigram\_following} respectively, to the \textbf{baseline} F1 and F2 GAM. This reduced the AIC by about 250 units for F1 and 80 units for F2 respectively. However, in the model for F1, the concurvity values for the smooth term for bigram\_following and bigram\_previous are 0.75 and 0.53 while the concurivty values for speaker and vowel sequence is much lower (0.05 and 0.12 respectively). For F2, the interaction of time and bigram\_following was not statistically significant ($p = 0.19$) and the concurvity values for the smooth terms are also high (0.75 for bigram\_following and 0.53 for bigram\_previous).

\parbox{0.48\textwidth}{
\begin{tabbing}
mm \= \texttt{pitch(logF1)} \= $\sim$ \= \texttt{gender} + \texttt{vowel} \kill
    \> \texttt{pitch(logF1)} \> $\sim$ \> \texttt{gender} +  \texttt{vowel} +\\
    \> \>\> s(\texttt{logF0}, by = vowel, k = 4) +  \\
    \> \>\> ti(\texttt{normalized\_time}, \texttt{logF0}) + \\
    \> \>\> s(\texttt{duration}, by = vowel, k = 4) + \\
    \> \>\> s(\texttt{normalized\_time}, \texttt{speaker}, bs = ``fs", m=1) + \\
    \> \>\> s(\texttt{utterance position}, by = vowel, k = 4) + \\
    \> \>\> s(\texttt{normalized\_time}, \texttt{vowel sequence} , bs = ``fs", m = 1) + \\
    \> \>\> \textbf{s(log\_bigram\_previous, k = 4)}+ \\
    \> \>\>\textbf{ti(normalized time, log\_bigram\_previous), k = c(4, 4)} + \\
    \> \>\> \textbf{s(log\_bigram\_following, k = 4)} + \\
    \> \>\> \textbf{ti(normalized time, log\_bigram\_following), k = c(4, 4)} +\\
    \> \>\> s(\texttt{normalized\_time}, \texttt{X}, bs = ``fs", m = 1) \\
\end{tabbing}
}

\parbox{0.48\textwidth}{
\begin{tabbing}
mm \= \texttt{pitch(logF2)} \= $\sim$ \= \texttt{gender} + \kill
    \> \texttt{pitch(logF2)} \> $\sim$ \> \texttt{gender} + \texttt{vowel} +\\
    \> \> \>s(\texttt{logF0}, by = vowel, k = 4) +  \\
    \> \> \>s(\texttt{duration}, by = vowel, k = 4) + \\
    \> \> \>s(\texttt{normalized\_time}, \texttt{speaker}, bs = ``fs", m=1) + \\
    \> \> \>s(\texttt{utterance position}, by=is.\textschwa, k = 4) + \\
    \> \> \>s(\texttt{normalized time}, \texttt{vowel sequence}, , bs = ``fs", m = 1) + \\
    \> \> \>\textbf{s(log\_bigram\_previous, k = 4)}+ \\
    \> \> \>\textbf{ti(normalized time, log\_bigram\_previous), k = c(4, 4)} + \\
    \> \> \>\textbf{s(log\_bigram\_following, k = 4)} + \\
    \> \> \>\textbf{ti(normalized time, log\_bigram\_following), k = c(4, 4)} +\\
    \> \> \>s(\texttt{normalized time}, \texttt{X}, bs = ``fs", m = 1) \\
\end{tabbing}
}

In cross validation, the sum of square error (SSE) of the GAM with \textbf{word} and \texttt{bigram\_probability} was undistinguishable  the GAM with \textbf{word} (for F1 $t_{(9)}=-3.68, p = 0.005$ and for F2 $t_{(9)}=0.99673, p = 0.345$). 

We also examined other key results, such as those related to LDL. Using the predicted pitch values from a GAM model with \textbf{word} as the core predictor and \texttt{bigram\_probability} as an additional control variable did not change the accuracy of LDL.  

Therefore, adding bigram\_probabilities as extra predictors into GAMs doesn't improve the prediction accuracy and doesn't change any established research, in order to avoid highly complex models, bigram\_probability is not included in as a control variables in current study.
\clearpage
\newpage
\bibliography{reference}

\end{CJK}
\end{CJK}
\end{document}